\documentclass{article}
\usepackage{spconf}
\usepackage{graphicx}
\usepackage{setspace}
\usepackage{amsmath}
\usepackage{dsfont}
\usepackage{amsfonts}
\usepackage{lipsum} 
\usepackage{multicol}
\usepackage{float}
\usepackage{bm}
\usepackage{color}
\usepackage{etoolbox}
\usepackage{soul}
\usepackage{amssymb}
\usepackage[linesnumbered,ruled,vlined]{algorithm2e}
\usepackage{mathtools,amssymb}
\usepackage{caption}
\usepackage{float}
\usepackage{amsthm}
\usepackage[caption = false,subrefformat=parens,labelformat=parens]{subfig}
\usepackage{adjustbox}
\usepackage{afterpage}
\usepackage{multirow}
\usepackage{mathrsfs}
\usepackage{booktabs}
\usepackage{url}

\title{Security-Preserving Federated Learning via \\Byzantine-Sensitive Triplet Distance}

\name{Youngjoon Lee, Sangwoo Park${}^\dagger$, Joonhyuk Kang
\thanks{This research was supported by the MSIT (Ministry of Science and ICT), Korea, under the ITRC (Information Technology Research Center) support program (IITP-2020-0-01787) supervised by the IITP (Institute of Information \& Communications Technology Planning \& Evaluation)}}
\address{KAIST, South Korea \\
${}^\dagger$King’s College London, United Kingdom
}

\begin{document}

\maketitle

\begin{abstract}
While being an effective framework of learning a shared model across multiple edge devices, federated learning (FL) is generally vulnerable to Byzantine attacks from adversarial edge devices. While existing works on FL mitigate such compromised devices by only aggregating a subset of the local models at the server side, they still cannot successfully ignore the outliers due to imprecise \emph{scoring} rule. In this paper, we propose an effective Byzantine-robust FL framework, namely \emph{dummy contrastive aggregation}, by defining a novel scoring function that sensitively discriminates whether the model has been poisoned or not. Key idea is to extract essential information from every local models along with the previous global model to define a distance measure in a manner similar to triplet loss. Numerical results validate the advantage of the proposed approach by showing improved performance as compared to the state-of-the-art Byzantine-resilient aggregation methods, e.g., Krum, Trimmed-mean, and Fang.
\end{abstract}
\section{Introduction}
Success of deep learning has been built upon massive utilization of data examples \cite{simeone2022machine} by summarizing the core information into deep neural networks. While traditional deep learning assumes availability of the entire data set at the central server side, such assumption becomes impractical when dealing with private data, e.g., medical data of the patients. Federated learning (FL) \cite{mcmahan2017communication, li2020federated} mitigates this privacy constraint by sharing the model parameter vector instead of data itself so that multiple distributed users (e.g., hospitals) can achieve a single shared AI model (e.g., disease predictor) as if it were trained with the whole data set. 
\begin{figure}
     \centering
     \includegraphics[width=\columnwidth]{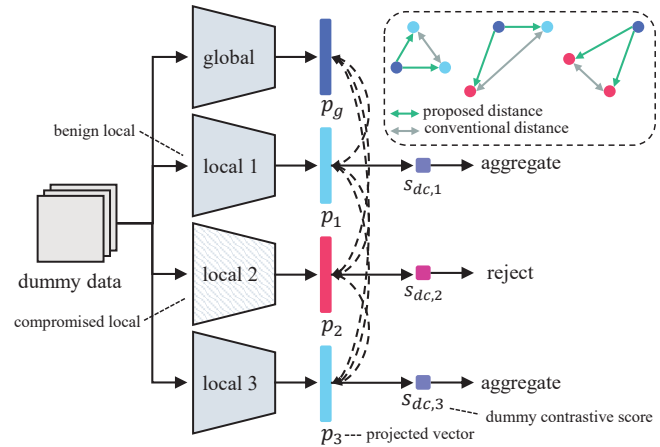}
    
    \caption{Illustration of the proposed \emph{dummy contrastive aggregation} for the design of secure FL. Based on the proposed triplet distance that uses the previous round's global model as an anchor, the proposed FL is robust to Byzantine attacks.}
     \label{fig:1}
     \vspace{-0.3cm}
\end{figure}

However, due to the nature of sharing the model parameter vector instead of data set, FL is generally more vulnerable to adversarial settings, which cannot apply robust learning techniques that mitigates outlier that works directly on the data set (see, e.g., \cite{zecchin2022robust} for robust loss design via $t$-logarithm). Accordingly, FL is susceptible to a variety of Byzantine failures, e.g., data poisoning \cite{tolpegin2020data} and model poisoning \cite{bhagoji2019analyzing}, \cite{cao2022mpaf} attacks, which leads to performance degradation \cite{bhagoji2019analyzing, bagdasaryan2020backdoor} and/or increased communication rounds \cite{li2019convergence}.

To overcome the backdoor attacks in FL, various kinds of aggregation rule at the central server side have been proposed, namely Krum \cite{blanchard2017machine}, Trimmed-mean \cite{yin2018byzantine}, and Fang \cite{fang2020local}. Essentially, these approaches mitigate the outliers by comparing the \emph{scores} of the shared model parameter vectors that are defined based on the pairwise distances between the shared local models.
However, as the scoring function is computed based only on the distance between the local models, the pairwise distance may lead to uninformative scoring function. To this end, we propose to also utilize the global model at the previous round, which can play as an \emph{anchor} that is likely to akin to the benign devices as compared to the compromised devices, to define a triplet distance as shown in Fig.~1. Furthermore, when computing the proposed triplet distance, instead of using the model parameter itself, we utilize feature vectors, or projected vectors \cite{li2021model,chen2020simple, chen2020big}, that is obtained by inferring some \emph{dummy data} to the model parameter vector of interest. In this paper, we generate dummy data following standard Gaussian distribution. 

\section{Preliminaries}\label{sec:Preliminaries}
In this section, we summarize existing Byzantine-Resilient aggregation strategies, namely Krum \cite{blanchard2017machine}, Trimmed-mean \cite{yin2018byzantine}, and Fang \cite{fang2020local}.
While Krum and Trimmed-mean work without any additional data set, Fang assumes availability of extra data set so as to further improve the performance of Krum or Trimmed-mean. In the following, we denote $\beta$ as the number of transmitted compromised edge devices known \emph{a priori} for aforementioned aggregation rules.

\textbf{Krum} \cite{blanchard2017machine}: Given $M$ locally updated models, Krum selects a single model for aggregation per each global epoch $g_e$. Precisely, Krum first computes the $\ell_2$-pairwise distance across every edge devices' models to compute the score of the $i$th model as 
\begin{equation}\label{beta}
s^{krum}_{\ell_2, i} = \sum_{ \theta^j\in \Im_{i,M-\beta-2}} ||\text{vec}(\theta^j)-\text{vec}(\theta^i)||^{2},
\end{equation}
where $\text{vec}(\cdot)$ denotes vectorization operation, and $\Im_{i,M-\beta-2}$ is the set of $M-\beta-2$ models that excludes $\beta-2$ models that are most apart from the $i$th model in $\ell_2$ distance manner. Then, the central server selects a single model which has the smallest score as
\begin{align}
    krum_\beta = \theta^{i_{krum}}\;\text{ for }\;i_{krum} = \arg \min_i s^{krum}_{\ell_2,i}.
\end{align}

\textbf{Trimmed-mean} \cite{yin2018byzantine}: Trimmed-mean removes the impact of outliers by taking the mean that excludes \emph{extreme} models. Similar to Kurm, Trimmed-mean first computes the $\ell_2$-pairwise distance scores as 
\begin{equation}\label{beta}
s^{trim}_{\ell_2, i} = \sum_{j=1}^M ||\text{vec}(\theta^j)-\text{vec}(\theta^i)||^{2}.
\end{equation}
Note that the score is computed with respect to all available pairs. Then the central server sorts the models in an ascending order based on the computed score $s_{\ell_2,i}$.
Once sorted, the central server removes the largest and smallest $\beta$ models and aggregate the remaining $M-2\beta$ models as
\begin{align}\label{trmean}
trmin_\beta = \frac{1}{M-2\beta} \sum_{ \theta^m\in \Im^{trim}_{M-2\beta}} \theta^m,
\end{align}
where $\Im^{trim}_{M-2\beta}$ is the set of $M-2\beta$ remaining models.
Note that the trimming parameter $\beta$ should be smaller than $M/2$.

\textbf{Fang:} \cite{fang2020local}: Unlike Krum or Trimmed mean, Fang utilizes the available global data $\mathcal{D}_g$ at the server-side to remove $\beta$ models. While Fang can be built upon both Krum and Trimmed mean, we focus here on Trimmed mean which has been reported to show better performance than application with Krum \cite{fang2020local}. In order to compute scoring function for $i$th model based on the available data set, Fang computes two aggregation models, one being the $trimin_\beta$ (4) with total $M$ models; while the other the $trimin_\beta$ (4) using $M-1$ models that excludes the $i$th model. We accordingly denote the corresponding Trimmed mean models as $A_i$ and $B_i$, respectively.
Then the central server computes the scoring function $s_{err, i}$ for each edge device $i$ as
\begin{equation}\label{Fang_2}
s_{err, i} = \mathcal{L}(A_i; \mathcal{D}_g) - \mathcal{L}(B_i; \mathcal{D}_g),
\end{equation}
where $\mathcal{L}(\theta;\mathcal{D})$ is the loss function of model $\theta$ using data set $\mathcal{D}$.
Then, the central server discards $\beta$ models which have the lowest scores $s_{err,i}$ to aggregate the $M-\beta$ remaining models $\Im^{fang}_{M-\beta}$ as
\begin{align}\label{fang}
fang_\beta = \frac{1}{M-\beta} \sum_{ \theta^m\in \Im^{fang}_{M-\beta}} \theta^m.
\end{align}

\section{System Model}\label{sec:model}
\subsection{Federated setting}
In this paper, we focus on the scenario of applying FL under targeted \cite{bhagoji2019analyzing} and untargeted \cite{cao2022mpaf} model poisoning attacks.
The federated learning network consists of $M$ edge devices including $B$ benign devices and $C$ compromised edge devices, communicating through the central server.
Each edge device $m=1,\ldots,M$ holds a different local dataset $\mathcal{D}_{m}$ that possibly contains a different number $|\mathcal{D}_{m}|$ of data points, i.e., $d_{m,1}, d_{m,2},...,d_{m,|\mathcal{D}_{m}|}$, with $i$th data example for $m$th device $d_{m,i}$.
The goal of FL is to train a globally shared model based on the edge devices' dataset $\{\mathcal{D}_{m}\}_{m=1}^M$ without sending data to the central server.
Mathematically, the training objective of FL can be written as $F(\theta) \triangleq \frac{1}{M} \sum_{m=1}^{M}f_{m}(\theta),$ where $F(\theta)$ is the global empirical loss over the entire edge devices' dataset $\{\mathcal{D}_m\}_{m=1}^M$, with the local empirical loss for edge device $m$ defined as $f_{m}(\theta)=\frac{1}{|\mathcal{D}_{m}|}\sum_{d_{m,i} \in \mathcal{D}_{m}}\mathcal{L}(\theta; d_{m,i}),$ denoting $\mathcal{L}(\theta; d_{n,i})$ as the local loss function for data $d_{n,i}$ computed from the model parameter $\theta$. In order to minimize the training objective $F(\theta)$, FL performs $G$ global epochs, in other words, $G$ communication rounds between edge devices and the central server.

\subsection{Benign model update}
At each global epoch $g_e$, every benign edge devices locally update their model with their own data based on the shared global model $\theta_{g_{e}}$ from the central server.
Precisely, each benign edge device $b \in B$ initializes $\theta_{g_{e}}$ to $\theta_{g_{e}-1}^{b, l_e=0}$ and trains its model $\theta_{g_{e}}^{b, l_e=0}$ up to local epoch $l_e=L$ with its own local data $\mathcal{D}_{b}$.
Assuming Stochastic Gradient Descent (SGD) optimization \cite{simeone2018brief}, at each local epoch $l_{e}\le L$, the local update rule can be written as
\begin{align}\label{local1}
\theta_{g_{e}}^{b,l_{e}+1} &\leftarrow \theta_{g_{e}}^{b,l_{e}} - \gamma\nabla_{\theta_{g_{e}}^{b,l_{e}}} \mathcal{L}_{ce}(\theta_{g_{e}}^{b,l_{e}};\tilde{D}_b) \\\nonumber
\theta_{g_{e}}^{b,l_{e}=0} &= \theta_{g_{e}},
\end{align}
with subset $\tilde{D}_b$ with $N$ examples, sampled from the local data set ${D}_b$. Here, $\gamma$ is the local learning rate and $\mathcal{L}_{ce}(\cdot)$ denotes the cross entropy loss \cite{simeone2018brief}.
Once this local training is finished, benign edge devices transmit their trained models $\theta_{g_{e}}^{b,l_e=L}$ to the central server.
Note that at the very first global epoch $g_{e}=0$, the global model parameter $\theta_{g_{e}=0}$ is randomly initialized.

\subsection{Compromised model update}
We now introduce targeted \cite{bhagoji2019analyzing} and untargeted \cite{cao2022mpaf} model poisoning updates for compromised edge devices. The goal of targeted model poisoning attack is to modify the global model's behavior on a small number of samples while maintaining the overall performance \cite{biggio2012poisoning}.
In contrast, the untargeted model only aims at degrading the performance of the global model \cite{bagdasaryan2020backdoor}, \cite{liu2017trojaning}. 

\textbf{Targeted attack} \cite{bhagoji2019analyzing}: In targeted attack,  compromised edge devices $c \in C$ perform additional step from \eqref{local1} as follows:
\begin{align}
\theta_{g_{e}}^{c,l_{e}=L} &\underset{attack}{\leftarrow} \theta_{g_{e}}^{c,l_{e}=L} + \delta_{g_{e}}^{c},\\\nonumber
\delta_{g_{e}}^{c} &= \lambda(\theta_{g_{e}}^{c,l_{e}=L}-\theta_{g_{e}}),
\end{align}
with boosting factor $\lambda$ that is designed to satisfy the compromised edge device's objective.

\textbf{Untargeted attack} \cite{cao2022mpaf}: In untargeted attack,  compromised edge devices $c \in C$ transmit the fake update without performing \eqref{local1} as follows:
\begin{align}
\theta_{g_{e}}^{c,l_{e}=L} &\underset{attack}{\leftarrow} \eta(\theta^\prime-\theta_{g_{e}}),\\\nonumber
\theta^\prime &\sim \mathcal{N}(0, I),
\end{align}
where $\eta$ is the scaling factor and $\mathcal{N}(0, I)$ denotes the standard multivariate Gaussian distribution.

\section{Dummy contrastive aggregation}\label{sec:dummy}
We now present the proposed aggregation method that is designed to alleviate the compromised devices introduce above. First key idea is to define a new scoring function in the projected vector domain unlike Krum or Trimmed mean. Since projected vector requires some input data to obtain the feature vector, we proposed to consider a dummy image as input for computing the scores.

At first, the central server randomly generates $N$ dummy data  $\{\xi_n\}_{n=1}^N$ where each $n$th sample has its element generated from the standard normal Gaussian $\mathcal{N}(0, 1)$. We set the size of each dummy input $\xi_n$ to be same as the input of the true data example $d_{m,i}$. With the generated dummy data set $\{\xi_n\}_{n=1}^N$, the projected vector $p$ of the $M$ local models and the global model are defined as
\begin{align}\label{dummy1}
p_m = g_{\theta_{g_{e}}^{m,l_{e}=L}}(\{\xi_n\}_{n=1}^N) \in \mathbb{R}^{N \times O},\\\nonumber
p_g = g_{\theta_{g_{e}-1}}(\{\xi_n\}_{n=1}^N) \in \mathbb{R}^{N \times O},
\end{align}
where $g_{\theta}(\cdot)$ denotes the neural network functionality before the last fully connected (FC) layer. Here, $O$ is the dimension of the projected vector for each dummy input $\xi_n$, which is determined by the neural network architecture.
Then the central server computes the dummy contrastive score $s_{dc,m}$ for each received model $\theta_{g_{e}}^{m, l_{e}=L}$ as follows:
\begin{equation}\label{dummy2}
s_{dc, i} = \sum_{j=1}^M \big(\mathcal{L}_{bce}(p_g;p_j)+\mathcal{L}_{bce}(p_g;p_i)\big),
\end{equation}
with the loss $\mathcal{L}_{bce}$ defined as 
\begin{equation}\label{dummy3}
L_{bce}(x;y)=\frac{1}{O}\sum_{o=1}^O -((y_o \cdot \log\sigma(x_o)+(1-y_o) \cdot \log(1-\sigma(x_o))),
\end{equation}
with sigmoid function $\sigma(\cdot)$. Note that in \eqref{dummy3}, we didn't consider $y_o$ as the probability measure to directly use the unnormalized logit vector $p_m$. Finally, the central server sorts the models in an ascending order according to $s_{dc,m}$ with respect to the other models.
After sorting, the central server removes the largest $\beta$ models and aggregate the remaining $M-\beta$ models as follows:
\begin{equation}\label{relativity}
\theta_{g_{e}+1} \leftarrow  \frac{1}{M-\beta} \sum_{ \theta^m\in \Im^{cont}_{M-\beta}} \theta^m,
\end{equation}
where $\Im^{cont}_{M-\beta}$ is the set of $M-\beta$ remaining models.
After all, $\theta_{g_{e}+1}$ is used for the next global epoch $g_{e}$ and we repeat the procedure until $g_{e}$ reaches the predefined value $G$.

\section{Experiment and Results}\label{sec:experiment}
\subsection{Experiment setting}
We use blood cell images dataset \cite{acevedo2020dataset} with ResNet-18 \cite{he2016deep} classifier.
We compare the proposed aggregation rule with Krum, Trimmed-mean, and Fang which are the commonly used methods in FL under adversarial setting. 
The considered blood cell microscope dataset consists of training set of 11,959 examples, validation set of 1,712 examples and a test set of 3,421 examples. We set the dimension of the projected vector $O$ as 512.
Additionally, we adopt \emph{quantity skew} \cite{li2022federated} over $M=10$ edge devices to take into account for non iid FL setting.
Other hyperparameter setting is available at the open source code\footnote{https://github.com/yjlee22/byzantineFL}.

\subsection{Results}
\subsubsection{Without Byzantine attacks.}
\begin{figure}
     \centering
     \includegraphics[width=0.8\columnwidth]{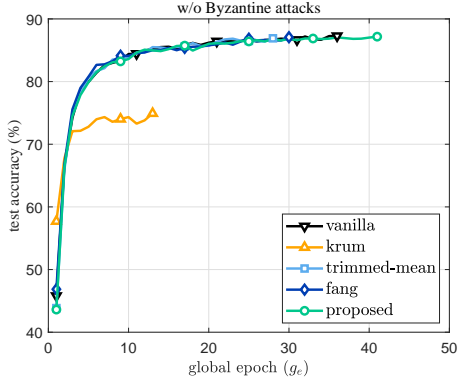}
     \caption{Test accuracy with respect to global epoch $g_e$ without any Byzantine attacks.}
     \label{fig:2}
\end{figure}

First, we check whether the proposed robust scheme works well enough as vanilla FL scheme, FedAvg \cite{mcmahan2017communication}, when there is no byzantine attacks.  To check the convergence of proposed method, we examine the test accuracy with respect to global epoch $g_e$. According to Fig. \ref{fig:2}, we can observe that the global model trained by the proposed method converge as well as vanilla FedAvg.
Therefore, it can be confirmed that the proposed approach can achieve approximately the same performance as the vanilla aggregation with a slight decrease in the rate of convergence.
\subsubsection{Impact of Byzantine percentage}
\begin{figure}
     \centering
     \includegraphics[width=\columnwidth]{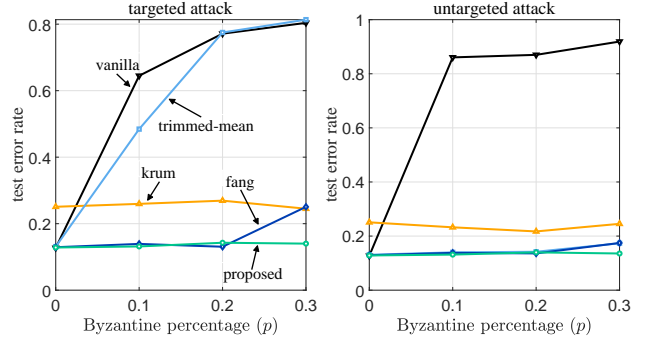}
     \caption{Minimum test error rate with respect to the percentage of compromised (model poisoning attacks) edge devices $p$ after sufficient round $G$ of communication rounds.}
     \label{fig:3}
\end{figure}

To check the backdoor attack effect, we examine the minimum test error rate after sufficient global epochs, as a function of the ratio of compromised edge devices $p := C/M$.
Additionally, we set $\beta$ to $C$ which is a key parameter for Krum, Trimmed-mean, and Fang.
According to Fig. \ref{fig:3}, unlike conventional approaches, it is shown that the proposed method works well regardless of the type of model poisoning attack. The proposed method outperforms all the schemes, while reaching the similar performance as compared to Fang, which requires additional global data. Note that our scheme does not require any extra data.

\subsubsection{Impact of non-iid degree}
\begin{figure}
     \centering
     \includegraphics[width=\columnwidth]{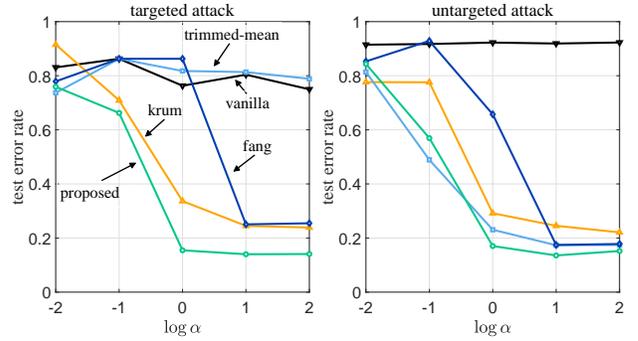}
     \caption{Minimum test error rate with respect to the non-iid degree $\alpha$ after sufficient round $G$ of communication rounds in the presence of Byzantine attacks (model poisoning).}
     \label{fig:4}
\end{figure}

To investigate the impact of non-iid degree $\alpha$ over Byzantine failures, we now compare the minimum test error rate with respect to $\alpha$ in a logarithmic scale.
Note that, the data distribution of edge devices becomes closer to iid as $\alpha$ increases.
In this experiment, the ratio of compromised devices is fixed to $p=0.3$.
From Fig. \ref{fig:4}, we can observe that the test error rate of proposed method decreases faster than the other Byzantine-resilient methods as $\alpha$ increases.
Unlike the proposed method and krum, we can also see that other techniques have different trends depending on the type of Byzantine attacks.
Therefore, it can be confirmed that the proposed method shows a consistent tendency regardless of the type of Byzantine attack and non-iid degree.
From this result, we believe that proposed method is more robust to backdoor attacks than existing methods in real-world problems.

\section{Conclusions}\label{sec:conclusion}
In this paper, we proposed dummy contrastive aggregation via alleviating Byzantine attacks which degrades the performance of FL.
Numerical results verify that the proposed approach outperforms the existing byzantine FL techniques, Krum, Trimmed-mean, and Fang, under the blood cell classification dataset, by simply changing the distance measure that is more sensitive to outliers. Future work may consider meta-learning \cite{park2020learning, park2020meta} for designing a dummy input to further improve the performance of FL with increased robustness to Byzantine attacks.

\bibliographystyle{IEEEbib}
\bibliography{reference}

\end{document}